\documentclass[letterpaper, 10 pt, conference]{ieeeconf}

\usepackage{times}
\usepackage{cite}
\usepackage{multicol}
\usepackage[pagebackref=false, letterpaper=true,colorlinks,bookmarks=true] {hyperref}
\usepackage{booktabs}
\usepackage{graphicx}
\usepackage{amsmath}
\usepackage{subcaption}
\usepackage{balance}
\usepackage{cprotect}

\IEEEoverridecommandlockouts
\begin{document}

\title{\LARGE \bf Integrating Disambiguation and User Preferences into \\ Large Language Models for Robot Motion Planning}

\author{Mohammed Abugurain and Shinkyu Park%
\thanks{
    The work was supported by funding from King Abdullah University of Science and Technology (KAUST), and the SDAIA-KAUST Center of Excellence in Data Science and Artificial Intelligence (SDAIA-KAUST AI).   } 
    \thanks{Abugurain and Park are with  Electrical and Computer Engineering Program, Computer, Electrical, and Mathematical Science and Engineering Division, King Abdullah University of Science and Technology (KAUST). \{\tt {mohammed.abugurain, shinkyu.park\}@kaust.edu.sa}}
    }

\maketitle
\begin{abstract}
This paper presents a framework that can interpret humans' navigation commands containing temporal elements and directly translate their natural language instructions into robot motion planning. Central to our framework is utilizing Large Language Models (LLMs). To enhance the reliability of LLMs in the framework and improve user experience, we propose methods to resolve the ambiguity in natural language instructions and capture user preferences. The process begins with an ambiguity classifier, identifying potential uncertainties in the instructions. Ambiguous statements trigger a GPT-4-based mechanism that generates clarifying questions, incorporating user responses for disambiguation. Also, the framework assesses and records user preferences for non-ambiguous instructions, enhancing future interactions. The last part of this process is the translation of disambiguated instructions into a robot motion plan using Linear Temporal Logic. This paper details the development of this framework and the evaluation of its performance in various test scenarios. 
\end{abstract}

\section{Introduction}
Recent advancements in Large Language Models (LLMs) \cite{DBLP:conf/naacl/DevlinCLT19, openai2023gpt4, NEURIPS2020_1457c0d6,NEURIPS2022_b1efde53, chowdhery2022palm, touvron2023llama, bommasani2022opportunities} demonstrate their remarkable capabilities, exemplified by their ability to process and generate human-like text. These advanced computational models have revolutionized various domains, including robotics, where LLMs are increasingly employed to bridge the communication gap between humans and machines. This allows robots to execute tasks based on complex human instructions. 
However, as LLMs are widely used in robotics applications, an increasing number of challenges arise in accurately interpreting humans' natural language commands that are ambiguous or contain their personal preferences, which are unknown to the robots.

Our research is motivated by growing concerns about \textit{hallucination} and \textit{inconsistency} in LLMs as reported in recent literature \cite{mallen-etal-2023-trust, tian-etal-2023-just, zhang2023language}.
Hallucination refers to generating false or misleading information that doesn't accurately reflect the input provided.  This is particularly problematic in robotic applications, as LLMs can misinterpret humans' instructions or generate unsafe commands when they hallucinate, ultimately impeding the applicability of LLM-integrated systems. Additionally, the inconsistency, referring to the phenomenon of LLMs producing different responses to the same query, further complicates the use of LLMs in scenarios where reliability of the robotic systems is crucial. These issues limit the functional scope of robots and raise concerns about safety, especially in environments where human-robot interaction is frequent. In this work, we address these limitations by enhancing the reliability of LLMs for robot motion planning and execution. We propose a framework that  effectively detects and corrects misinterpretation and inconsistencies, thereby bridging the gap between the potential of LLMs and their practical, reliable application in robotics.

Our framework, which can be integrated into existing robotic systems using LLMs, minimizes ambiguity in human natural language instructions, enabling robots to understand and execute the instructions with improved reliability. This is challenging because human natural language is inherently ambiguous, where meanings are frequently implied rather than explicitly stated. For instance, given an instruction \textit{``Go to the room where I forgot my keys''}, the robot must understand where the user forgot their keys to plan and execute the navigation command.
Furthermore, the robot needs to determine whether a given human instruction contains all the information about their preferences required to execute the instructed task. To illustrate, consider an instruction \textit{``Go to the room I like the most''}. The instruction appears unambiguous as there is exactly one room the user likes the most, but it does not indicate which room they like the most.
In this example, the robot must detect that the instruction is incomplete and ask the user for more information about their preference. 

Drawing from these two examples, we consider ambiguity as the uncertainty in a sentence's meaning that allows for multiple, equally probable interpretations. Instructions with unspecified preferences, though not ambiguous, are regarded as incomplete, necessitating additional information on the preferences to complete the task. This paper posits that ambiguity is a universal phenomenon, not specific to any particular user, while preference is user-specific. 

As our main contributions, we develop a framework capable of resolving ambiguity in natural language instructions and clarifying them through meaningful queries. Additionally, it can detect whether the instructions contain all the necessary information regarding user preferences, query the users whenever additional details are required to complete the task, and also retain this information for future use.
The rest of the paper is organized as follows: We present the background and related work in Section \ref{sec:related_work}, and the problem formulation and proposed framework in Section \ref{sec:problem_formulation}. We validate the efficacy of our framework through experiments in Section~\ref{sec:results}, and conclude the paper with summaries and discussions in Section~\ref{sec:conclusion}.

\begin{figure*}[ht]
	\centering\includegraphics[width=1\textwidth]{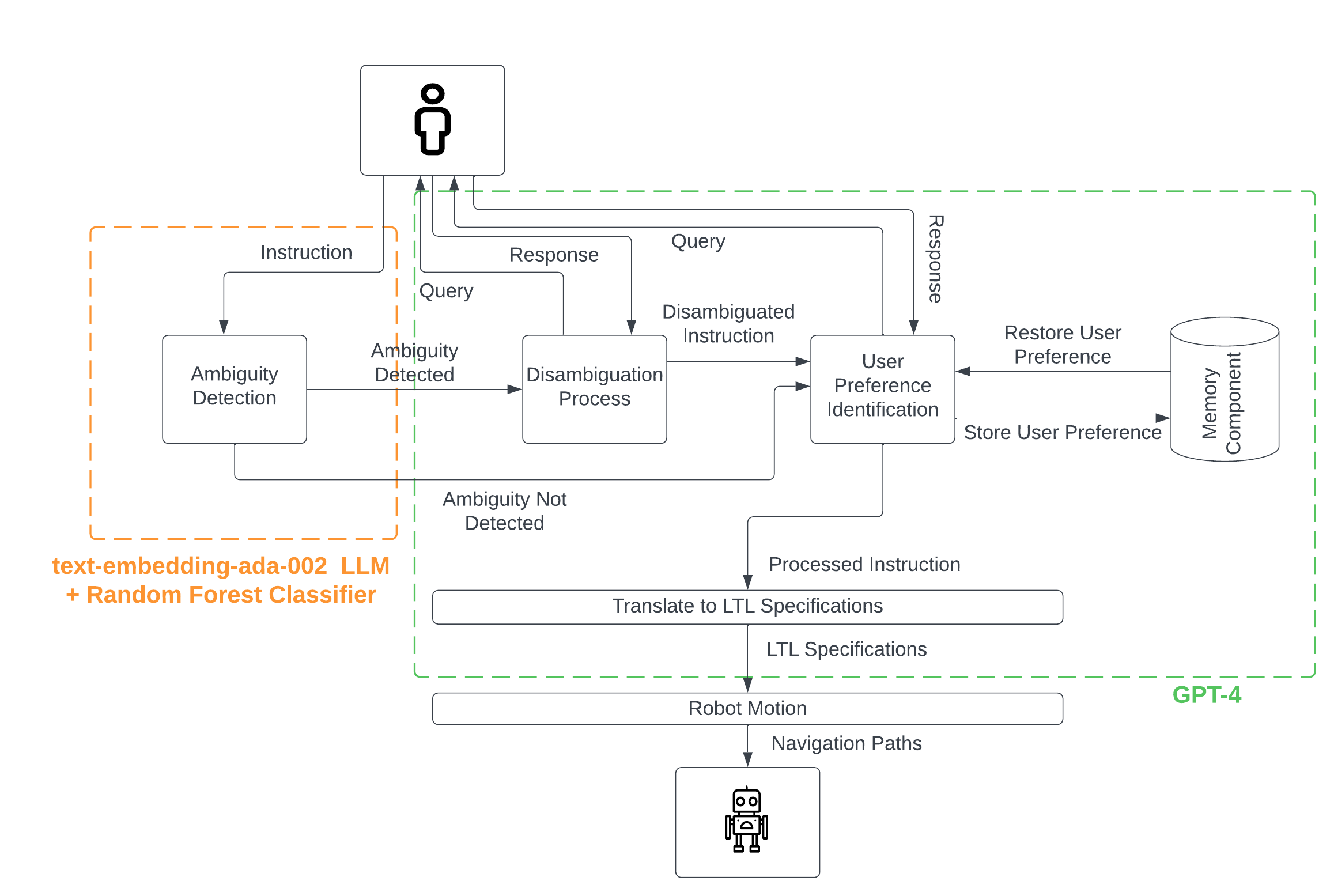}
    \cprotect\caption{Our proposed framework consisting of \verb|text-embedding-ada-002| with a random forest classifier for the ambiguity detection in human instructions, and GPT-4 for the disambiguation process and user preference identification. The framework integrates existing methods for translating the processed instructions into LTL specifications and for planning navigation paths based on these specifications.}
 \label{fig:framework}
\end{figure*}

\section{Related Work}
\label{sec:related_work}
\textit{a) Instructing Robots by Natural Language:} This is appealing because the end user does not need to learn technical skills to operate the robot, ensuring smoother and more accessible experience. The work of \cite{lynch2021language} integrates free-form natural language into imitation learning and robot perception into a single neural network model. This approach enabled users to instruct robots on their manipulation tasks using natural language.
The work of \cite{DBLP:conf/iser/DuvalletWHHOTRS14} introduces a probabilistic framework to instruct robots to navigate unknown environments using natural language instructions. With advancements in LLMs, recent works have utilized LLMs to translate human natural language into robot commands \cite{huang_instruct2act_2023, lin_text2motion_2023, 10203853, ahn_as_2022,  huang_language_2022}. The work of \cite{huang_instruct2act_2023} provides a Python-based interface to integrate a LLM, allowing a robot to
retrieve information about its surrounding environment and manipulate objects. Based on natural language instructions from users, the LLM generates a sequence of Python functions for the robot to execute in order to complete the required object manipulation task.
The work of \cite{lin_text2motion_2023} presents a natural language-based robot planning framework for multi-step sequential manipulation tasks. This framework constructs task and policy plans to achieve symbolic goals as instructed by a user.

\textit{b) Resolving Ambiguity in Natural Language:} A substantial body of recent work investigates methods to reduce the ambiguity in natural language instructions for robotics applications. 
To mention a few, the work of \cite{park2023clara} utilizes LLMs for evaluating the clarity of user commands for human-robot interactions, with the goal of enhancing the reliability of robot operations by minimizing unintended actions. Using LLMs with with the map of the environment, they introduce a method to classify commands as clear, ambiguous, or infeasible. For ambiguous commands, a question-generation technique is employed to seek clarification. 


The work of \cite{majumdar2020improving} develops a framework designed to enhance the efficacy of LLMs in multi-step planning and commonsense reasoning within robotics applications. This framework improves the LLMs' ability to recognize their own uncertainty, prompting them to seek human inputs as necessary. It incorporates conformal prediction theory to provide statistical guarantees for a robot's completion of instructed tasks with minimal human intervention. Similarly, the work of \cite{dogan_asking_2022} introduces a framework that enables robots to request clarifications in conversations with human users regarding ambiguous object identifications. This approach aims to improve efficiency and user satisfaction by leveraging the knowledge about the robot's surrounding environment to reduce the frequency of such requests, thereby minimizing the need for human intervention.

Though we share a similar motivation with previous studies \cite{park2023clara, majumdar2020improving, dogan_asking_2022} to disambiguate human instructions, our work stands apart by focusing on robot navigation for which we design a framework that directly detects ambiguity in the instructions.
The framework is required to translate human natural language instructions into LTL specifications, necessitating the development of a tailored framework. Consequently, as shown in Table~\ref{tab:ambiguity_analysis}, our framework outperforms the method of \cite{park2023clara} in terms of ambiguity detection accuracy within the targeted applications. Additionally, our framework integrates the retention of information about user preferences through a proper querying process, which can be utilized in future interactions.

\textit{c) User Preference Identification and Retention:} This is a crucial step toward enabling robots to understand and adapt to individual user preferences. Such adaptability is essential for enhancing the user experience and building trust between humans and robots. The work of \cite{doi:10.1080/09540090802413145} proposes a decision-theoretic model based on a POMDP formulation to improve human-robot interaction. This approach enables robots to learn user preferences in real time, despite ambiguity and noise in their speech, thereby reducing the need for offline model training. The model's effectiveness was demonstrated through simulations and experiments in a robotic wheelchair application. More recent work \cite{wu2023tidybot} focuses on implementing a robot-assisted household cleanup system in which LLMs for few-shot summarization are utilized to learn user preferences from a few examples. These preferences include the designated places for certain household objects.

\section{Problem Formulation and Methods}
\label{sec:problem_formulation}
Consider an environment containing $m$ locations of interest, denoted as $L = \{l_1, l_2, \hdots, l_m\}$, each assigned a unique label recognizable by a robot, as illustrated in Fig.~\ref{fig:experiment}. We investigate the problem of planning a path $P = \{p_1, p_2, \hdots, p_n\}$ for robot navigation. The start point $p_1 \in L$, all the intermediate locations $\{p_2, \cdots, p_{n-1}\} \subset L$, and the goal point $p_n \in L$ are to be selected according to human instructions provided in natural language, where we assume that the instructions can be specified by LTL. When the user instructions are clear, existing approaches allow for translating natural language instructions into LTL specifications \cite{chen_nl2tl_2023,pan_data-efficient_2023,mavrogiannis_cook2ltl_2023} and facilitate robot motion planning with LTL specifications \cite{kurtz2023temporal}.
Therefore, we aim to develop a framework that removes ambiguity and identifies user preferences in the instructions. Fig.~\ref{fig:conversation} illustrates a dialog between a user and a robot during the disambiguation and user preference identification process within our framework. A robot navigation path resulting from the dialog is illustrated in Fig.~\ref{fig:experiment}.


We utilize the \verb|text-embedding-ada-002| model \cite{openai_2024_new_embeddings} and a random forest classifier to analyze ambiguity in the instructions. 
\verb|text-embedding-ada-002| is a LLM trained on a large text corpus. It analyzes and converts textual input into high-dimensional vectors, capturing intricate semantic and contextual nuances, which can be used for various natural language processing applications, including text classification \cite{10391005}. In our framework, the output of \verb|text-embedding-ada-002| is fed into a trained random forest classifier to determine whether a sentence is ambiguous. The random forest model is beneficial as it can classify the vector data produced by \verb|text-embedding-ada-002| without the need for dimensionality reduction. 

When ambiguity is detected in the instructions, our framework employs the GPT-4 model to generate contextually relevant questions to resolve it. These questions aim to solicit additional information from the user, effectively narrowing down multiple interpretations of the instructions. The same GPT-4 model then processes the user's responses to produce disambiguated, explicit instructions.
Upon removing ambiguity from the original instructions, the framework attempts to detect any unspecified user preferences using GPT-4. We equipped the GPT-4 model with a conversation buffer memory to record users' preferences for future interactions. If the preference is known from previous interactions stored in the memory component, the model will not ask the user for their preference again. Hence, this step is crucial for personalized interactions and enhancing the user experience. 

The final step involves translating the processed instructions into LTL specifications. LTL is a formalism used to specify a sequence of events alongside their temporal interrelations. Such specification is crucial in accurately representing scenarios where the order in which a robot's actions are executed is critical, such as in robotic task planning and navigation. LTL's core strength emanates from its ability to represent complex temporal scenarios within a structured, mathematically rigorous framework. This capability facilitates the precise interpretation and execution of sequences by machines, ensuring high levels of accuracy and predictability. To leverage these compelling attributes, we incorporated LTL representation into the design of our framework.

We adopt the framework developed in \cite{chen_nl2tl_2023,pan_data-efficient_2023,mavrogiannis_cook2ltl_2023} to translate the user instructions into LTL specifications. We have integrated this framework into our framework based on GPT-4.
Then, the framework computes paths for robot navigation achieving the LTL specifications, using the approach proposed in  \cite{kurtz2023temporal}. This approach represents the path planning problem as searching for the shortest path over a graph of convex sets, which can be solved using convex optimization.



Through this multifaceted approach, our system not only addresses the challenges posed by the ambiguity of natural language but also enriches the interaction between humans and robots by understanding and adapting to individual user preferences. Fig.~\ref{fig:framework} illustrates our framework.

\section{Experiments}
\label{sec:results}
\subsection{Dataset and Training}
\label{sec:dataset_traning}
To optimize the performance of the framework, we train the random forest classifier. We begin with creating a dataset that comprises of 155 ambiguous statements of navigation commands. Each statement is accompanied by a map of the environment, a clarification question, possible valid user responses, and a disambiguated sentence for each valid response. The dataset is generated through the following process: Initially, we generate 20 ambiguous statements of navigation commands across three different environments. Then, we use GPT-4 to paraphrase and manipulate the word order in these statements to produce additional 135 ambiguous statements.\footnote{The dataset and implementation code will be available after the paper review.} Then, we combined our dataset with existing ones from the previous works \cite{Gopalan-RSS-18, oh19}. The dataset from \cite{Gopalan-RSS-18} consists of 3382 commands corresponding to 39 LTL statements for mobile robot manipulation and pick-and-place tasks.\footnote{The authors from \cite{oh19} used Amazon Mechanical Turk to collect a dataset of 810 natural language instructions and corresponding 27 LTL expressions. Each collected data point is then expanded by augmenting other atomic proposition relevant to the environment, resulting in 6185 commands corresponding to 343 LTL expressions.}
Since the dataset is unbalanced, with the number of ambiguous statements being less than that of non-ambiguous statements, we performed upsampling to balance the dataset. Subsequently, 20 ambiguous and 20 unambiguous statements were selected for testing. 

Using the dataset, we train the random forest classifier with the standard training configuration of $100$ trees in the forest. The input to the classifier is the vector representations of human instructions produced by \verb|text-embedding-ada-002|. The classifier then outputs 0 for non-ambiguous and 1 for ambiguous statements. 
 
\subsection{Ambiguity Detection and Disambiguation Process}
The first step in our framework is to analyze the ambiguity of input sentences and classify the sentences as ambiguous or non-ambiguous.
We compared the performance of our ambiguity detection model, which consists of \verb|text-embedding-ada-002| combined with a random forest classifier, against that of a range of existing models that could be applied to our problem.
These models include \verb|text-embedding-ada-002|  with a support vector machine, zero-shot GPT-3.5, zero-shot GPT-4, few-shot GPT-3.5, few-shot GPT-4, BERT, and the CLARA framework proposed in \cite{park2023clara}. As summarized in Table~\ref{tab:ambiguity_analysis}, the \verb|text-embedding-ada-002| model with a random forest classifier outperformed the others in terms of accuracy.

\begin{table}[ht]
	\caption{Performance Comparison of Various LLM-based Ambiguity Detection Methods}
	\label{tab:ambiguity_analysis}
	\begin{center}
		\begin{tabular}{cc}
			\toprule
			\textbf{Model}               & \textbf{Accuracy}  \\
			\midrule\midrule
			\textbf{ADA + Random Forest}          & \textbf{0.85}                 \\
			\midrule
			Few-shot GPT-4               & 0.79                     \\
			\midrule
			BERT                         & 0.71                \\
			\midrule
			Zero-shot GPT-4              & 0.69                  \\
			\midrule
			Few-shot GPT-3               & 0.61                    \\
            \midrule
            CLARA \cite{park2023clara} & 0.53 \\
			\midrule
			ADA + Support Vector Machine \hspace{3em} & 0.51                       \\
			\midrule
			Zero-shot GPT-3              & 0.48                       \\
			\bottomrule
		\end{tabular}
	\end{center}
\end{table}

\begin{table*}[ht!]
\centering
	\caption{Examples of the Disambiguation Process}
	\label{tab:disambiguation}
	\footnotesize 
	\begin{tabular}{p{0.25\textwidth}p{0.23\textwidth}p{0.18\textwidth}p{0.23\textwidth}}
		\toprule
		Ambiguous Sentence                                                                                         & Clarification Question                                                                                                                                 & User Response                                          & Disambiguated Sentence                                                                              \\
		\midrule
		Go to the room and then pass by the red room                                                                & Which room are you referring to when you say "go to the room"?                                                                                         & the green room                                         & Go to the green room and then pass by the red room.                                                 \\
		\midrule
		Go to the room and then pass by the red room                                                                & Which room should I go to first before passing by the red room?                                                                                        & the blue room                                          & Go to the blue room first and then pass by the red room.                                            \\
		\midrule
		Make your way to the room that feels less crowded                                                           & Which room are you referring to as the one that feels less crowded?                                                                                    & the red room                                           & Make your way to the red room that feels less crowded.                                              \\
		\midrule
		Go to the room where we had the meeting.                                                                    & Which room are you referring to when you mentioned "the room where we had the meeting"?                                                                & green room                                             & Go to the green room where we had the meeting.                                                      \\
		\midrule
		Head to the room, and then go past the crimson chamber.                                                     & Which room are you referring to when you say "the room"? Additionally, could you confirm if the "crimson chamber" refers to the 'red room' on the map? & the room is the blue, and the crimson is the green one & Head to the blue room, and then go past the green chamber.                                          \\
		\midrule
		Once you've bypassed the red room, navigate through the blue room to find your way to the illuminated space & To clarify your command, could you please specify what you mean by the "illuminated space"?                                                            & red room                                               & Once you've bypassed the red room, navigate through the blue room to find your way to the red room. \\
		\bottomrule
	\end{tabular}
\end{table*}

\begin{figure*}[h!]
	\centering\includegraphics[width=1\textwidth]{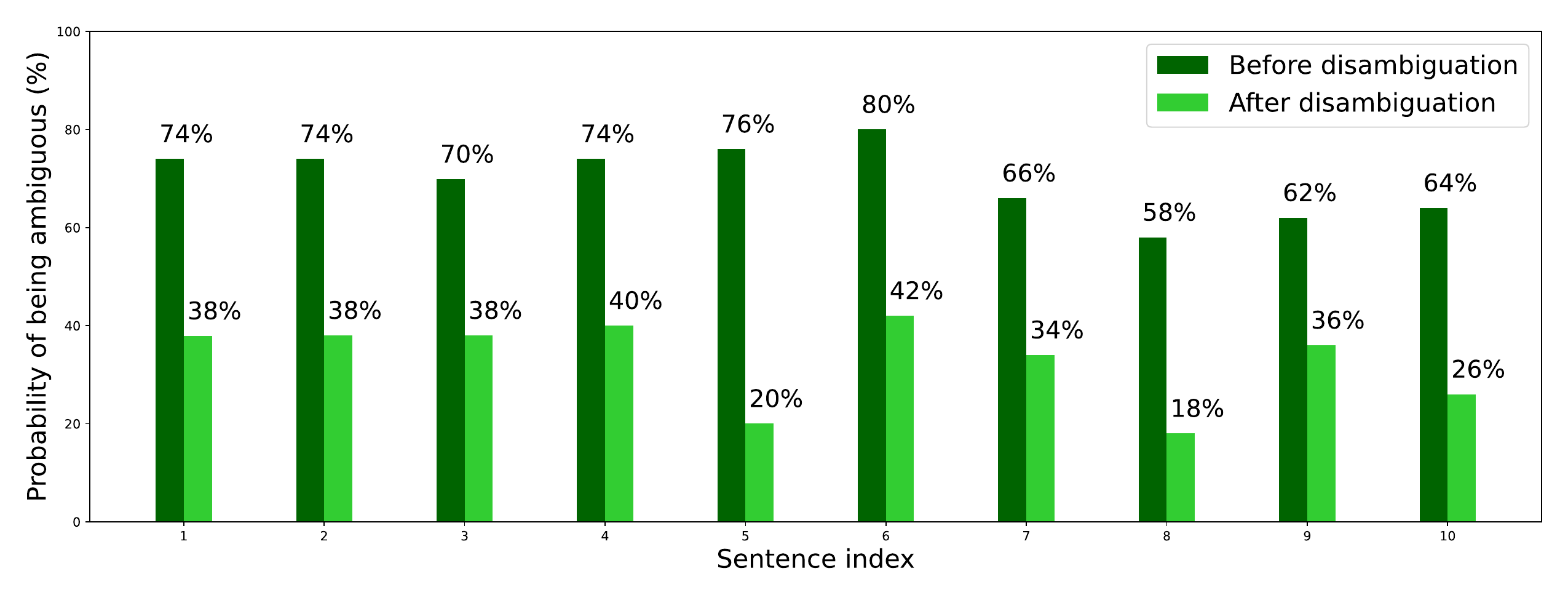}
	\caption{Ambiguity likelihood before and after the disambiguation process}
 \label{fig:disambiguation}
\end{figure*}

When the input sentence is classified as ambiguous, the framework generates questions to resolve the ambiguity. The questions are generated using GPT-4 by providing the input sentence and the environment map. The questions are then presented to the user, and their responses are fed into  GPT-4 to generate a resulting disambiguated sentence. Table \ref{tab:disambiguation} lists a few examples of original ambiguous sentences, clarifying questions, user responses, and resulting disambiguated sentences.

To assess the performance of the disambiguation process, we conducted a series of tests using the dataset we created. These tests involved a comparative analysis of the likelihood of sentences being ambiguous before and after the disambiguation process. Initially, sentences were passed through the ambiguity classifier to determine their likelihood of being ambiguous. Then, ambiguous sentences were refined in our disambiguation process based on user responses. The refined sentences were then re-analyzed by the classifier to assess any changes in their ambiguity likelihood. This approach allowed us to quantitatively measure the effectiveness of the disambiguation process.
The results are shown in the bar graph in Fig.~\ref{fig:disambiguation}, 
which illustrates that the disambiguation process effectively reduces the likelihood of ambiguity by $36.8\%$ on average.

\subsection{Preference Analysis}
The next step in our framework involves preference analysis using the GPT-4 model. For this purpose, we design a prompt for the model to process the input sentence, the environment map, and the previous interactions from the memory component. When a user preference is detected in the input sentence, the model will first check its memory component to see if the user preference is already known. If not, the model will generate a question to ask the user for their preference. The user's responses are then fed into the model to generate a sentence that reflects the user's preference. Finally, the model records the user's preference in its memory component.


\subsection{Application to Robot Planning}

\begin{figure*}
    \centering
    \includegraphics[width=1\textwidth]{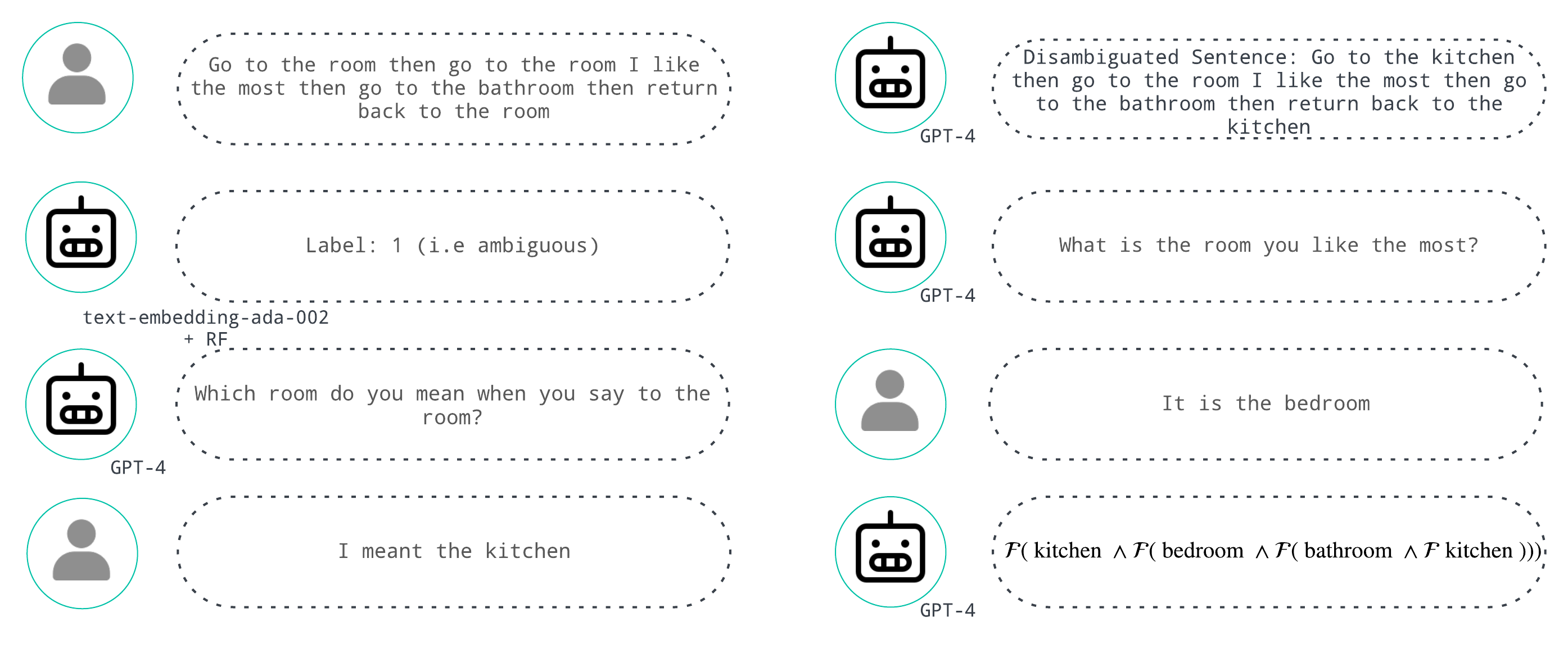}
    \caption{Dialog between a user and a robot in the disambiguation and user preference identification process.}
    \label{fig:conversation}
\end{figure*}

\begin{figure*}[ht]
    \centering
    \begin{subfigure}{0.16\textwidth}
        \includegraphics[width=\linewidth]{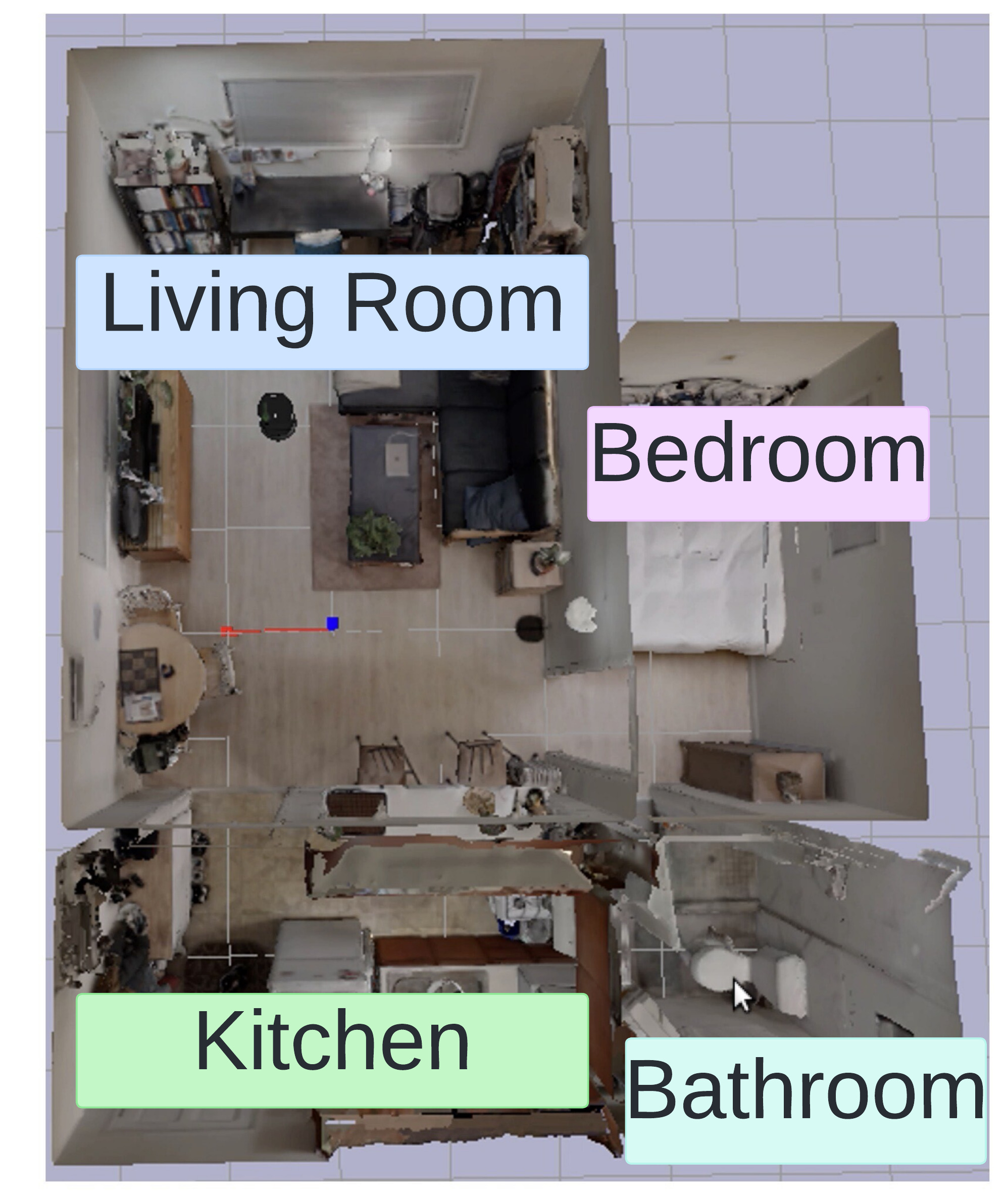}
        \caption{}
        \label{fig:sub1}
    \end{subfigure}
    \hfill
    \begin{subfigure}{0.16\textwidth}
        \includegraphics[width=\linewidth]{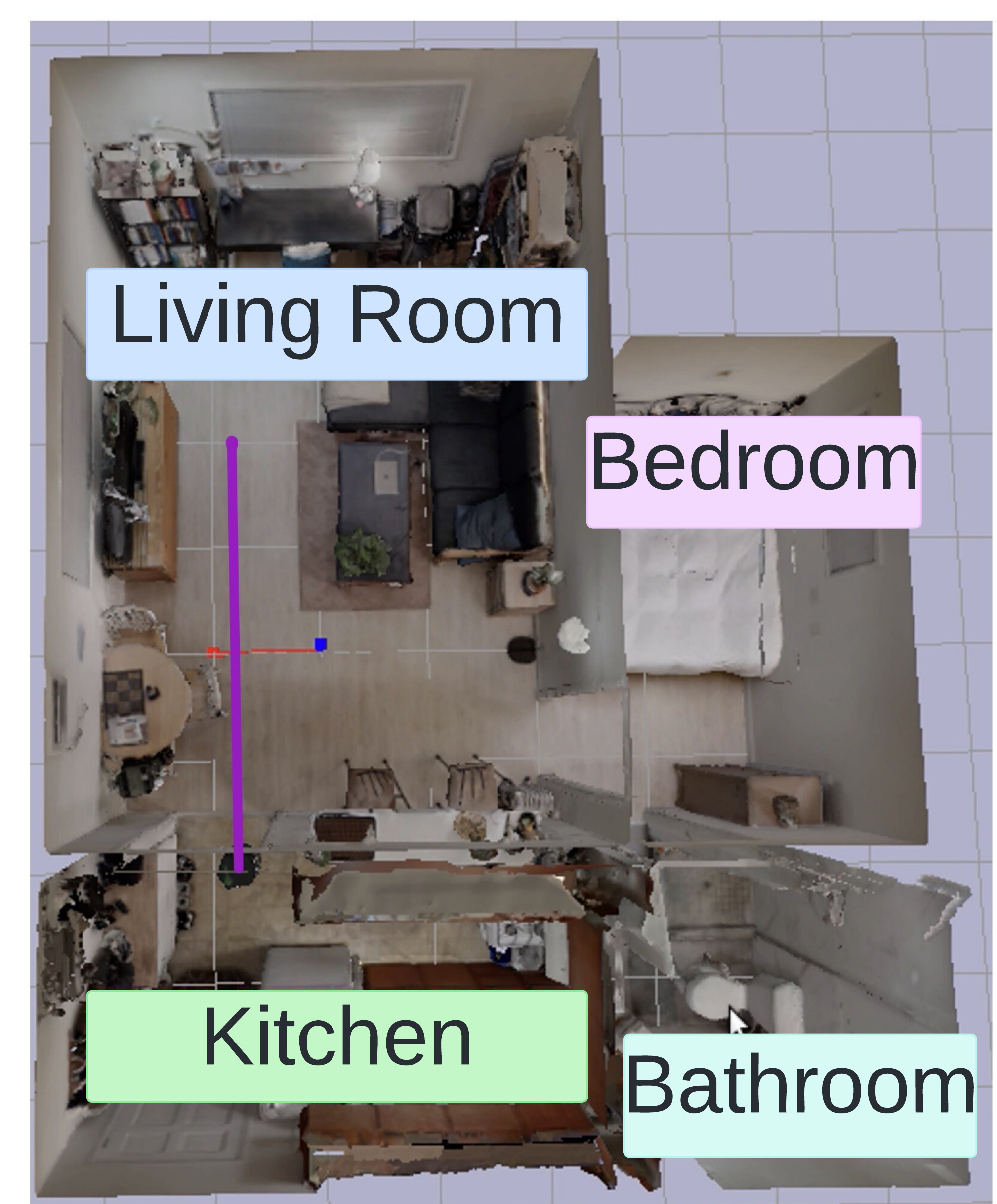}
        \caption{}
        \label{fig:sub2}
    \end{subfigure}
    \hfill
    \begin{subfigure}{0.16\textwidth}
        \includegraphics[width=\linewidth]{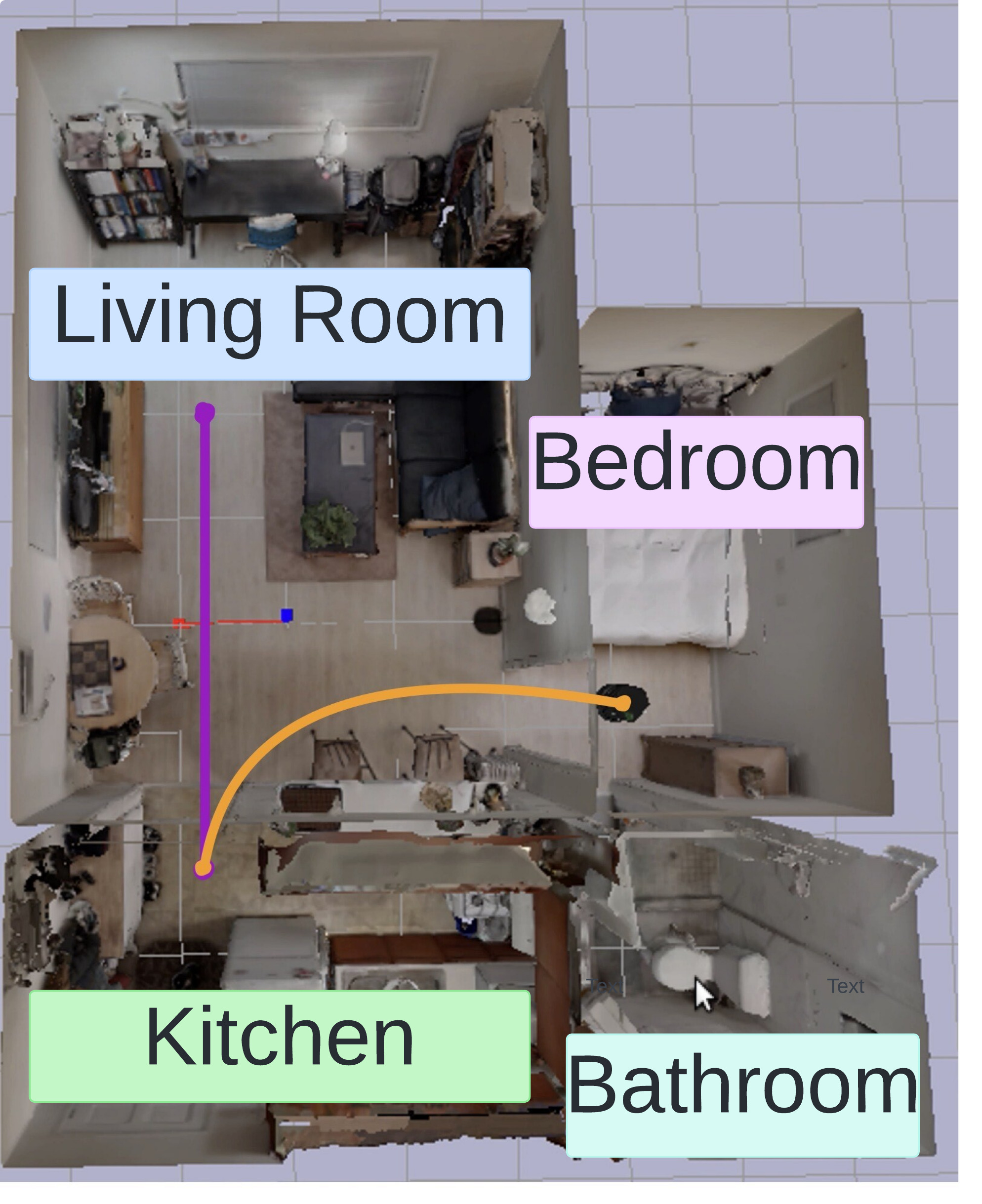}
        \caption{}
        \label{fig:sub3}
    \end{subfigure}
    \hfill
    \begin{subfigure}{0.16\textwidth}
        \includegraphics[width=\linewidth]{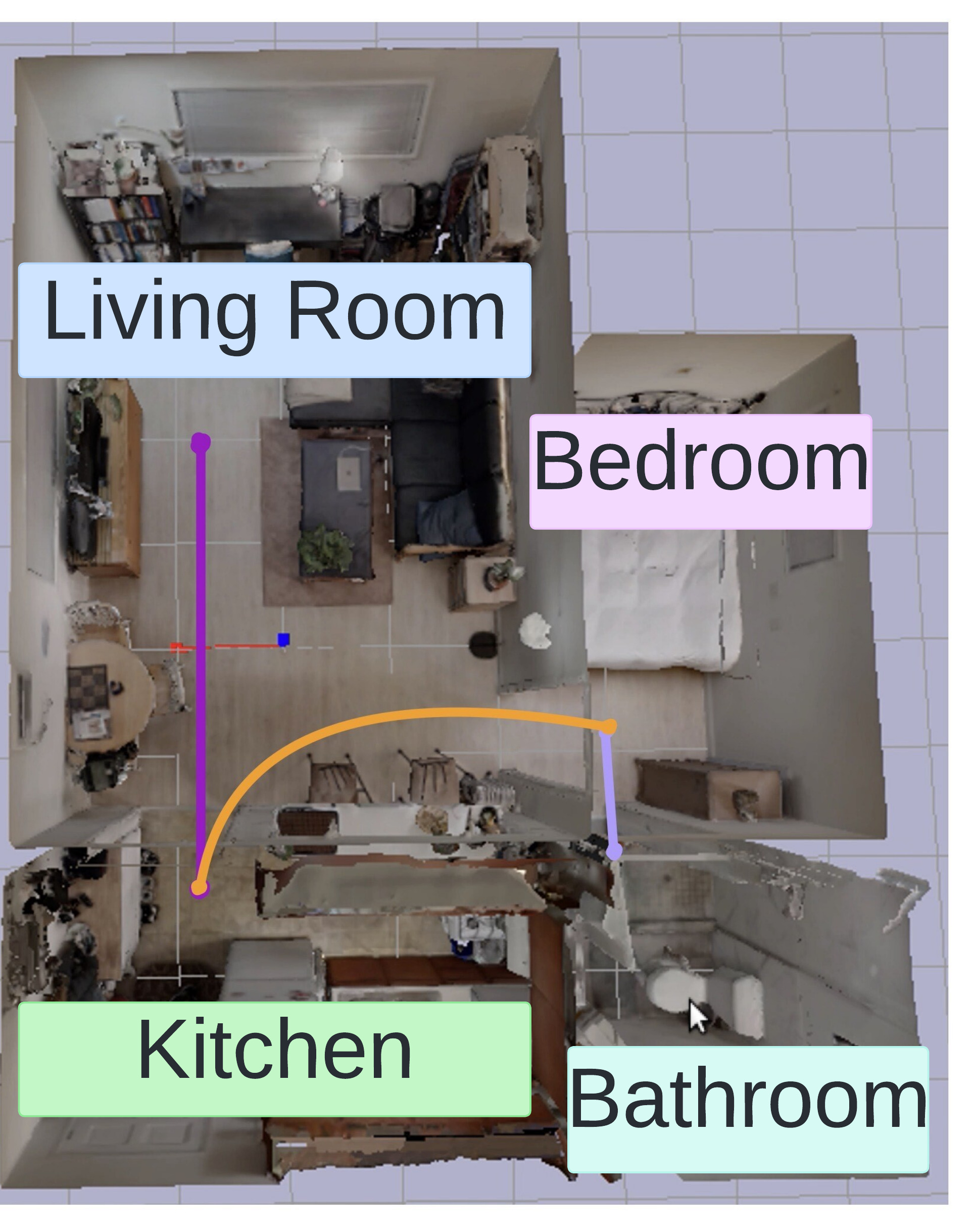}
        \caption{}
        \label{fig:sub4}
    \end{subfigure}
    \hfill
    \begin{subfigure}{0.16\textwidth}
        \includegraphics[width=\linewidth]{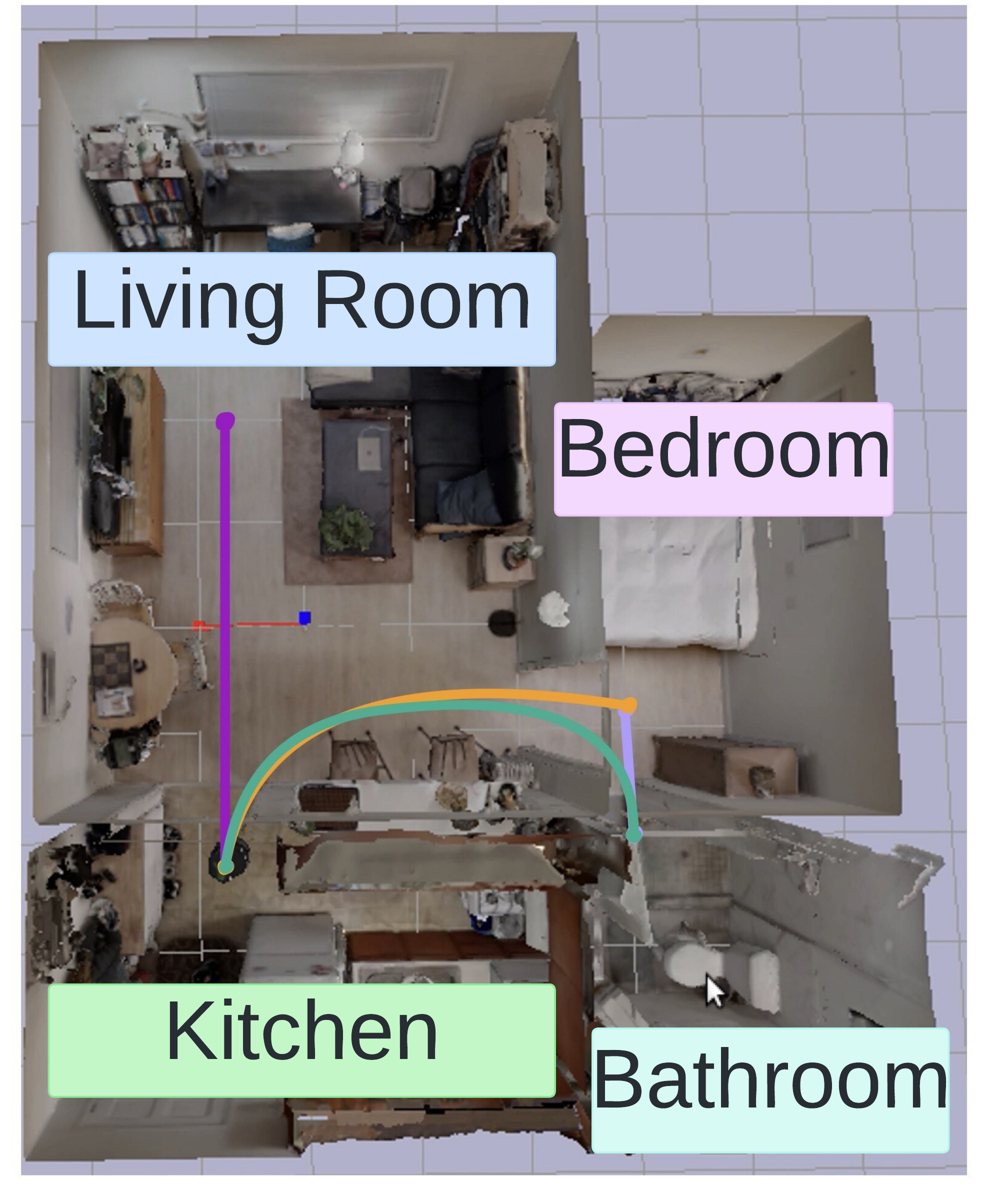}
        \caption{}
        \label{fig:sub5}
    \end{subfigure}
    \caption{Validation of the proposed framework for robot navigation: Each figure illustrates a distinct stage in the robot's navigation as it follows the user's instructions (refer to Fig.~\ref{fig:conversation} for the instructions.)}
    \label{fig:experiment}
\end{figure*}

The dialog depicted in Fig.~\ref{fig:conversation} provides an example of how our framework can be used to translate user navigation instructions into their corresponding LTL specifications. In our example, the user provides an ambiguous instruction, which includes an unspecified personal preference, for a robot to sequentially visit multiple locations in the environment. When the dialog is initiated between the robot and the user, the framework disambiguates the instruction and identifies the user's preference, and then it translates the processed instruction into the corresponding LTL specifications. We adopt the algorithm in \cite{kurtz2023temporal} to compute paths for the robot navigation based on the specifications. The robot's navigation along the computed paths is illustrated in Fig.~\ref{fig:experiment} where we can verify that the robot is able to precisely carry out the user's instructions. This verifies the efficacy of our framework.


\section{Conclusion}
\label{sec:conclusion}

We have developed an LLM-based framework that can translate a user's navigation instructions into paths that enable the robot to carry out the instructions precisely. To enhance the reliability of the LLM in translation, we have proposed methods to detect and address ambiguity and unknown user preferences in the instructions. \verb|text-embedding-ada-002| combined with a random forest classifier was optimized for ambiguity detection, and GPT-4 was utilized to solicit user inputs to clarify ambiguity and unknown preferences. To improve the user experience, we added a memory component to GPT-4 to retain information about the user's preferences for future interactions. By integrating with existing methods for translating the processed instructions into LTL specifications and computing robot navigation paths that meet these specifications, our framework offers an end-to-end capability to convert natural language instructions into viable navigation paths. Through performance comparisons with other models in ambiguity detection and validation in a robot navigation scenario, we have verified the effectiveness of our framework. Future research will explore integrating visual input to refine the robot's understanding of complex instructions and environments. Moreover, another direction will be integrating real-time feedback for dynamic path adjustment and expanding language support for broader applicability.



\section*{Acknowledgments}
\label{sec:acknowledgments}
The authors would like to thank Mujtaba Al Hubayl for his assistance in developing the figures and comments on the paper.

\balance
\bibliographystyle{IEEEtranBST/IEEEtran.bst}
\bibliography{Bib/modified_references}

\end{document}